\documentclass[letterpaper]{article} 
\usepackage{aaai25}  
\usepackage{times}  
\usepackage{helvet}  
\usepackage{courier}  
\usepackage[hyphens]{url}  
\usepackage{graphicx} 
\usepackage{natbib}  
\usepackage{caption} 
\usepackage{amsmath}
\usepackage{amssymb}
\usepackage{colortbl}
\usepackage{subcaption}
\usepackage{booktabs} 
\usepackage{multirow}
\usepackage[misc]{ifsym}

\usepackage[dvipsnames, svgnames, x11names]{xcolor}
\definecolor{lgray}{gray}{0.7}

\usepackage{algorithm}
\usepackage{algorithmic}
\usepackage{enumitem}

\usepackage{newfloat}
\usepackage{listings}
\DeclareCaptionStyle{ruled}{labelfont=normalfont,labelsep=colon,strut=off} 
\floatstyle{ruled}
\newfloat{listing}{tb}{lst}{}
\floatname{listing}{Listing}
\nocopyright

\title{\nds{}: Enhanced Adaptive Spatial-wise Sampling and View-wise Fusion Strategies for NeRF-based Indoor Multi-view 3D Object Detection}
 \author {
    Chi Huang \textsuperscript{\rm 1},
    Xinyang Li\textsuperscript{\rm 1},
    Yansong Qu\textsuperscript{\rm 1},
    Changli Wu\textsuperscript{\rm 1},
    Xiaofan Li\textsuperscript{\rm 2},
    Shengchuan Zhang\textsuperscript{\rm 1} $^{(\textrm{\Letter})}$,
    Liujuan Cao\textsuperscript{\rm 1}
}

\affiliations {
    \textsuperscript{\rm 1}Key Laboratory of Multimedia Trusted Perception and Efficient Computing, Ministry of Education of China, Xiamen University, 361005, P.R. China.\\
    \textsuperscript{\rm 2}Baidu Inc.\\
    huangchi@stu.xmu.edu.cn, imlixinyang@gmail.com, quyans@stu.xmu.edu.cn, wuchangli@stu.xmu.edu.cn, shalfunnn@gmail.com, zsc\_2016@xmu.edu.cn, caoliujuan@xmu.edu.cn
}

\usepackage{bibentry}

\newcommand{\nds}[1]{\textit{NeRF-DetS}}

\def\√{\checkmark} 
\def\eg{\emph{e.g.}}

\begin{document}

\maketitle

\begin{abstract}

    In indoor scenes, the diverse distribution of object locations and scales makes the visual 3D perception task a big challenge. 
    Previous works (\eg, NeRF-Det) have demonstrated that implicit representation has the capacity to benefit the visual 3D perception task in indoor scenes with high amount of overlap between input images.
    However, previous works cannot fully utilize the advancement of implicit representation because of fixed sampling and simple multi-view feature fusion. 
    In this paper, inspired by sparse fashion method (\eg, DETR3D), we propose a simple yet effective method, \nds{}, to address above issues. \nds{} includes two modules: \textit{Progressive Adaptive Sampling Strategy} (\textit{PASS}) and \textit{Depth-Guided Simplified Multi-Head Attention Fusion} (\textit{DS-MHA}). 
    Specifically,
    (1)~\textit{PASS} can automatically sample features of each layer within a dense 3D detector, using offsets predicted by the previous layer.
    (2)~\textit{DS-MHA} can not only efficiently fuse multi-view features with strong occlusion awareness but also reduce computational cost.
    Extensive experiments on ScanNetV2 dataset demonstrate our \nds{} outperforms NeRF-Det, by achieving +5.02\% and +5.92\% improvement in mAP under IoU25 and IoU50, respectively. Also, \nds{} shows consistent improvements on ARKITScenes.
    
\end{abstract}
%

\section{Introduction}
3D object detection is a crucial task for extracting semantic and geometric information in 3D space, which is essential for applications such as augmented reality (AR), virtual reality (VR), and autonomous driving, among others. The primary objective is to identify and locate objects belonging to predefined classes within a 3D scene by processing data collected from various sensors.
Classical 3D object detection methods, including VoxelNet\cite{pointnet}, PointNet\cite{pointnet}, and VoteNet\cite{Votenet}, typically rely on point clouds acquired through 3D sensors such as LiDAR or radar. However, these methods entail high sensor deployment cost, and the available training data obtained through 3D sensors is limited. 
Therefore, perceiving and extracting 3D information from traditional RGB images, which is more cost-effective, has become a meaningful yet more challenging field. Exploring the relationship between posed RGB images in 2D space and various potential semantic information in 3D space has emerged as a significant research objective. This paper focuses on investigating how to fully utilize features and information from multi-view posed images to implement 3D object detection in static indoor scenes.
\begin{figure}
    \centering
    \includegraphics[width=0.45\textwidth] {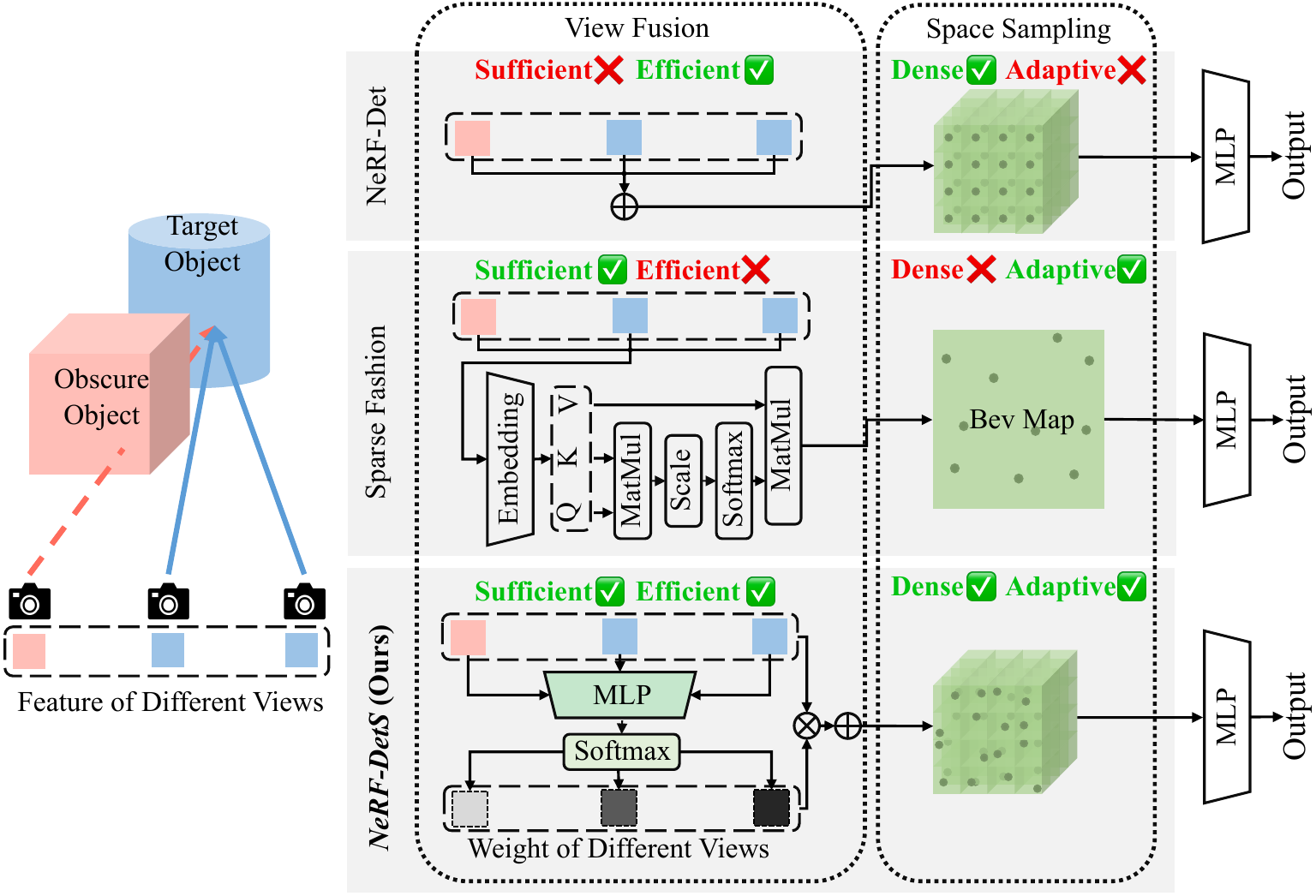}
    \caption{\textbf{Comparison between \nds{} and other methods}. There are two main steps in 3D object detection using 2D images: view fusion and space sampling strategies. We compare our method with NeRF-Det and the Sparse Fashion method.} 
    \label{fig:main}
\end{figure}
As a novel 3D representation method, Neural Radiance Field (NeRF) introduces volumetric rendering into the process of 3D reconstruction. 
By utilizing different views of posed images, NeRF is capable of reconstructing high-quality 3D scenes. 
NeRF-based approaches, such as Plenoxels \cite{plenoxels} and IBRNet \cite{ibrnet}, have been applied in spatial perception methods to enhance the ability of perception models to understand 3D space.
Specifically, NeRF-Det \cite{NeRF-Det} proposes a novel perception pipeline that utilizes NeRF as an intermediate representation. By leveraging the spatial geometric information obtained from the NeRF branch, it explicitly models the opacity field of the 3D space.
This dual-branch approach, using NeRF as an intermediate representation, ensures the continuity of features in space. Moreover, the presence of the NeRF branch enables self-supervision of information across different views, ensuring the correctness and consistency of features extracted from multi-view images.
Although NeRF-Det has demonstrated the effectiveness of the structure that combines perception with the NeRF branch, NeRF-Det does not fully leverage NeRF's strengths in detection tasks. In particular, NeRF's ability to introduce feature continuity in the spatial domain has not been fully exploited.
Additionally, NeRF-Det fails to effectively fuse features from different viewpoints during the fusion of multi-view features at the same sampling point.
For a sampling point in space, occlusion occurs across multiple viewpoints. Simply averaging the features sampled from different viewpoints introduces significant noise and fails to fully capture the characteristics of the sampling points in space.

Currently, in the field of autonomous driving, one popular approach is to adopt methods similar to DETR3D \cite{detr3d}, which are based on a sparse and multi-layer sampling strategy, as well as attention mechanism.
Inspired by this, we introduce an improved adaptive sampling method based on dense sampling, the \textit{Progressive Adaptive Sampling Strategy} (\textit{PASS}) to replace the original fixed sampling method. This approach fully leverages the advantages of spatial feature continuity and multi-view feature consistency.
In our approach, the prediction from the previous layer includes an offset to guide the sampling in the subsequent layer. Compared to fixed sampling, this strategy allows for supplementing the information missed at previous sampling points. 
Additionally, considering the dense sampling strategy in indoor scenes, we propose a new multiple viewpoints fusion method: the \textit{Depth-Guided Simplified Multi-Head Attention Fusion} (\textit{DS-MHA}). 
This method accounts for the distance between sampling points and feature planes (i.e., depth) by predicting the multi-head weights and corresponding multi-head features of multiple viewpoints for the each sampling point in space.
This approach enables the sufficient and efficient fusion of features from different views in space with a reasonable computational cost.

Extensive experiments demonstrate that by using the proposed method \nds{}, the results achieved under the current dual-branch detection pipeline approach show significant improvement compared to previous works.

Our key contributions are summarized as follows:
\begin{enumerate}
  \item We employ the \textit{PASS} to adaptively sample points in space, supplementing spatial information missed by previous layers.
  \item Introducing the \textit{DS-MHA} approach, we effectively integrate features from sampling points across multiple perspectives with reduced computation cost.
  \item Our method outperforms NeRF-Det on ScanNetV2 dataset, by achieving +5.02\% and +5.92\% improvement in mAP under IoU25 and IoU50.
\end{enumerate}

\section{Related Works}

\noindent \textbf{3D Object Detection.} 
The progression of autonomous driving technology and the growing popularity of AR and VR devices have spurred rapid advancements in 3D object detection research. This is evident from the development of methods such as BevFusion \cite{bevfusion}, Camliflow \cite{camliflow} and GAPretrain \cite{huang2023geometric}.
Early 3D detection methods were generally adapted from 2D, like FCOS3D \cite{fcos3d}.
Nowadays, mature outdoor 3D detection methods focus on BEV (bird's-eye view) perception, such as BEVFormer \cite{bevformer}, BEVDet \cite{bevdet}, SparseBEV \cite{sparsebev} and VCD \cite{huang2024leveraging}. Subsequently, many occupancy reconstruction approaches like TPVFormer \cite{tpvformer} and VoxFormer \cite{voxformer} have also been developed. DETR3D \cite{detr3d}  and PETR \cite{petr} lift 2D features from multi-views into 3D to conduct 3D object detection based on DETR \cite{DETR} and Deformable DETR \cite{deformable}. Additionally, numerous methods have emerged for indoor detection task, including RGB-D scan-based methods such as 3D-SIS \cite{3dsis} and point cloud-based approaches like VoteNet \cite{Votenet} and FCAF3D \cite{FCAF3D}. ImVoxelNet \cite{imvoxelnet} utilizes images from multiple viewpoints to extract 2D features, which are subsequently projected from 2D to 3D, thereby representing the 3D space as a voxel for perception.	

\noindent \textbf{Neural Radiance Fields.} 3D representation has gained significant attention since the vanilla NeRF \cite{nerf}. Early works, such as DVGO \cite{dvgo}, NSVF \cite{nsvf}, Mip-NeRF \cite{mipnerf}, used differentiable volumetric rendering to fit geometric and RGB information in 3D space with a network. 
NeuS \cite{neus} and VolSDF \cite{volsdf}  improve the reconstruction quality by using SDF to replace density for better surface geometry reconstruction.
Subsequent methods, Plenoxels \cite{plenoxels} and 3DGS \cite{Gaussian}, have also demonstrated significant success in explicitly storing color and density features within the spatial domain. Instant-NGP \cite{INGP} adopted a multi-resolution hash encoding as storage. These approaches have proven highly effective in leveraging multi-view posed images for robust 3D space reconstruction and generating high-quality novel view synthesis (NVS). 
Many works, such as MVS-NeRF \cite{mvsnerf}, IBRNet \cite{ibrnet}, PixelNeRF \cite{pixelnerf}, and RegNeRF \cite{regnerf}, have also explored how to synthesize new views using a limited number of posed images.
In particular, methods such as IBRNet \cite{ibrnet} and PixelNeRF \cite{pixelnerf} project 2D image features into 3D space and decode the 3D features into the geometry and RGB information of positions in the 3D space.	
Inspired by this line of thought, NeRF-Det \cite{NeRF-Det} projects ray samples onto the image plane and demonstrates the effectiveness of this method.	



\noindent \textbf{Perception with NeRF.} Numerous works NSVF \cite{nsvf}, Panoptic NeRF \cite{panopticnerf}, NeRF-RPN \cite{NeRF-RPN}, PeRFception \cite{PeRFception}, NeRF-Det \cite{NeRF-Det}, NeRF-Det++ \cite{NeRF-Detpp} and NeRF-MAE \cite{nerfmae} have already incorporated NeRF into all kinds of perception tasks. In the context of 3D detection tasks, NeRF-RPN \cite{NeRF-RPN} and PeRFception \cite{PeRFception} first construct the scene into a NeRF representation and then perform perception on it. However, this approach requires reconstructing the entire scene in space before conducting perception, making it a non-end-to-end process that consumes considerable time. Additionally, some information may be lost during the reconstruction process. Recently, several works, such as NeRF-Det \cite{NeRF-Det} and NeRF-Det++ \cite{NeRF-Detpp}, focus on improving perception performance with NeRF, demonstrating its effectiveness. 
With the growing maturity of NeRF in 3D representation and deeper exploration of its role in perception, NeRF-Det \cite{NeRF-Det} is the first to incorporate a NeRF branch into the pipeline for 3D object detection.	
The NeRF branch serves as an additional branch, using a geometry MLP to predict the opacity of the sampling points.	
The opacity information helps the detection head focus on areas with high opacity.	
Furthermore, the NeRF branch fosters effective self-supervision among multi-view images.	
However, existing efforts have not sufficiently leveraged the advantages of introducing the NeRF branch, particularly the continuity of features across the entire scene.	
Therefore, we propose \nds{} with \textit{Progressive Adaptive Sampling Strategy} (\textit{PASS}) and \textit{Depth-Guided Simplified Multi-Head Attention Fusion} (\textit{DS-MHA}).
The \textit{PASS} fully utilizes the continuity of the feature in 3D space through adaptive sampling, rather than aggressively increasing the volume resolution of sampling points.
The \textit{DS-MHA} efficiently fuse multi-view features with strong occlusion awareness with reduced computational cost.

\begin{figure*}[ht]
    \centering
    \includegraphics[width=1\textwidth] {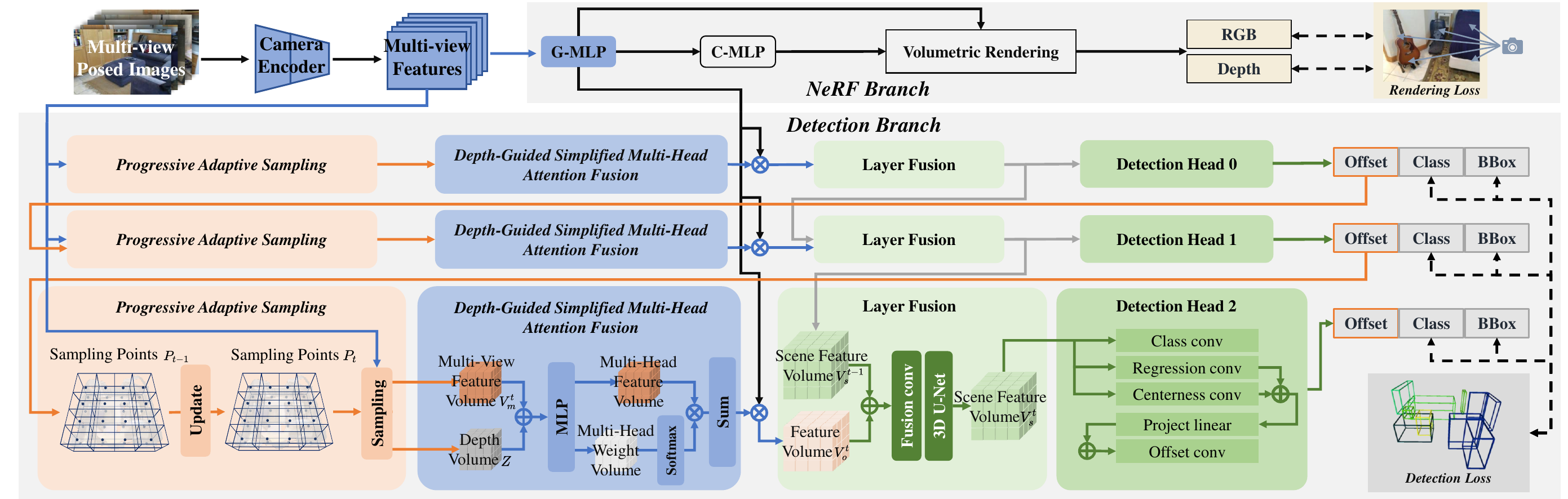}
    \caption{\textbf{Overview of \nds{}}. \textit{Progressive Adaptive Sampling Strategy} includes Progressive Adaptive Sampling and the Layer Fusion. Progressive Adaptive Sampling updates the sampling points for subsequent layer and Layer Fusion fuses the feature from the previous layer.
    The blue part is the process of \textit{Depth-Guided Simplified Multi-Head Attention Fusion}. The dashed lines represent our supervision. In the whole process, we not only use the geometry information from NeRF branch, but also use predicted offset of previous layer to guide the sampling process, and leverage depth to guide the fusion of features from multi-views.}
    \label{fig:overview}
\end{figure*}

\section{Preliminary}
 In this work, we present a novel method, \nds{}, which follows the main pipeline of NeRF-Det for 3D object detection. It takes only posed multi-view images as input and outputs a set of 3D bounding boxes that indicate the locations and classifications of objects in space. 

First, $N$ images are randomly selected from different views of the scene, and then passed through the camera encoder to extract their features. The output 2D features from multi-view images are denoted as $F_{i} \in \mathbb{R} ^ {w\times h \times c } $, where $i = 1,2,3,...,N $, and where $w$ and $h$ represent the size of the feature map, and $c$ is the feature dimension.

For a sampling point in 3D space with coordinate $\mathbf{p}=(x, y, z)^{T}$, and a corresponding camera parameters of viewpoint $i$, including intrinsic matrix $K_{i} \in \mathbb{R}^{3\times3}$ and the extrinsic matrix $R_{i} \in \mathbb{R}^{3\times4}$: $\left(u_i^{\prime},v_i^{\prime},d_i\right)^T=K_i\times R_i\times(\mathbf{p},1)^T$, where $
(u_i,v_i)=(u_i^{\prime}/d_i,v_i^{\prime}/d_i)$
We can obtain the pixel position $(u_i, v_i)$ of the point $\mathbf{p}$ when it is projected onto the viewpoint $i$'s camera plane.
Then, interpolation is used to query 3D points from the 2D features, $V_i(\mathbf{p})=\text{Interpolate}((u_i,v_i),F_i)$. This process allows us to obtain features of the sampling points from multiple viewpoints in the 3D space.
 
The NeRF branch include two parts: G-MLP and C-MLP. G-MLP outputs the density of the point $\mathbf{p}$, while C-MLP outputs the RGB of  $\mathbf{p}$ under the view direction $\mathbf{d}$:
\begin{equation}
\begin{aligned}
\sigma(\mathbf{p}),\mathbf{\hat{h}}(\mathbf{p})&=\text{G-MLP}(V(\mathbf{p}),\gamma(\mathbf{p})),\\ 
\mathbf{c}(\mathbf{p},\mathbf{d})&=\text{C-MLP}(\mathbf{\hat{h}}(\mathbf{p}),\mathbf{d}).
\end{aligned}
\end{equation}
Where $\gamma$ is the positional encoding. In the training process, the colors and depths of novel view cameras are rendered by using RGB and density of points sampled along the rays: 
\begin{equation}
\mathbf{\hat{C}}(r)=\sum_{i=1}^{N_p}T_i\alpha_i\mathbf{c}_i,D(r)=\sum_{i=1}^{N_p}T_i\alpha_it_i,
\end{equation}
where $T_{i}= \exp \left ( - \sum _{j= 1}^{i- 1}\sigma _{j}\delta _{t}\right ) $, $\alpha _{i}=$ $1- \exp \left ( - \sigma _{i}\delta _{t}\right ) $, $t_i$ is the distance between sampled $i$th point between the camera, $\delta_t$ is the distance between sampling points along the ray.
Simultaneously, G-MLP converts the density of sample points from the detection branch into opacity $\alpha(\mathbf{p})=1-\exp{(-\sigma(\mathbf{p}))}$. This provides spatial opacity information as a feature to the detection branch.

\section{Method}

\subsection{Overall Detection Process}
Following the pipeline of NeRF-Det, we get the 2D features from multi-view as s $F_{i} \in \mathbb{R} ^ {w\times h \times c } $, where $i = 1,2,3,...,N $.
Then, we uniformly sample spatial coordinates throughout the scene to obtain the original coordinate set as sampling points  $P_{0} \in \mathbb{R} ^ {W\times L \times H \times  3 }$. Here, $W$, $L$, and $H$ are the number of points sampled on different axes. After sampling the features in scene, we employ \textit{DS-MHA} to fuse the multi-view feature volume from  $ V_m \in\mathbb{R} ^ {N \times W\times L \times H \times  c }$ into the volume $ V_o \in\mathbb{R} ^ {W\times L \times H \times  c }$. After the feature volume $V_o$ passes through the 3D U-Net, we obtain $V_s \in\mathbb{R} ^ {W\times L \times H \times  c}$, representing the whole scene. $V_s$ is used to predict the bounding boxes, classifications, and offsets. 
Unlike NeRF-Det, which uniformly samples at static sampling points for each scene, our method uses predicted offset to update the sampling points from $P_t$ to $P_{t+1}$ in the subsequent layers.
For these layers, the updated sampling points are used to sample features in the scene.
After \textit{DS-MHA}, we fuse the feature volume $ V_o ^{t}$ with the previous scene feature volume $V_s^{t-1}$ during layer fusion to obtain $V_s^{t}$.
In this way, \textit{PASS} adaptively supplements the scene's features. 
The NeRF branch is used to render the color and depth of novel views, and the Geometry MLP within the NeRF branch generates the geometry volume $V_G$ for the sampling points. The geometry volume is used to make the detector geometry-aware following the \textit{DS-MHA} process.


\subsection{Progressive Adaptive Sampling Strategy}











Throughout the entire process, the perception branch ensures that the features extracted from images contain semantic information. The NeRF branch enhances the consistency of features extracted from multi-view through self-supervision, owing to NeRF's inherent characteristics. Additionally, the NeRF branch supervises the extraction of spatial features for novel views, ensuring the continuity of information in space. 
Furthermore, the opacity derived from Geometry MLP (G-MLP) in NeRF branch directs the focus of features toward important and relevant object positions, thereby improving perception performance.  In NeRF-Det, features are sampled at fixed sampling points:
\begin{equation}
V_s = \Theta(\text{Sample}({(F_i)}_{i=1}^N, P)    
\end{equation}
where $\Theta$ denotes the 3D U-Net, $F_i$ represents the multi-view image feature at the $i$th view and $P$ refers to fixed sampling points.

The continuity of the scene is a significant advantage of NeRF over other traditional reconstruction or NVS methods. However, NeRF-Det does not fully exploit this advantage. Therefore, we propose the \textit{Progressive Adaptive Sampling Strategy} (\textit{PASS}), which includes Progressive Adaptive Sampling and Layer Fusion as shown in Figure \ref{fig:overview}, to flexibly and efficiently utilize the continuous features in space to enhance perception in a dense sampling fashion. 
\begin{equation}
V_s^{t} = \Theta_t(V_s^{t-1}, \text{Sample}({(F_i)}_{i=1}^N, P_{0} + \Delta_{t-1}(V_s^{t-1}))  
\end{equation}
where $\Theta_t$ represents the Layer Fusion process, $V_s^{t -1}$ is the scene feature volume of previous layer, $P_0$ refers to original sampling points and $\Delta_{t-1}(V_s^{t-1})$ denotes the predicted offset of previous layer.

Specifically, at first, we uniformly sample at voxel coordinates as the original sampling points $P_0$. After sampling and fusion, we obtain a feature volume $V_{o}^0$. We pass it through a 3D U-Net to generate $V_{s}^0$, which represents the entire scene. In \textit{PASS}, the detection head not only to predicts the bounding boxes and classes of objects in the 3D space, like other FCOS3D-like methods do, but also the offset of the next layer sampling points relative to the original sampling points.
We concatenate the input scene features $V_s^0$ with the predicted results from the regression convolution layer and the centerness convolution layer after linear mapping. The concatenated features then undergo the offset convolution layer to predict offsets.
To obtain the samplinng points $P_1$ of the first layer, we use the original sampling points $P_0$ and the predicted offsets. 

Through the same sampling and fusion process, we obtain $V_{o}^1$, which represent the feature volume sampled at the first-layer sampling points $P_1$ after the offset. 
After concatenating $V_{o}^1$ with the feature volume $V_{s}^0$, we use a shallow fusion convolution layer and a 3D U-net with the same structure to generate feature volume of the first layer $V_{s}^1$. 
After obtaining the new feature volume  $V_{s}^1$, we follow the same process to derive $V_{s}^2$, by predicting the bounding boxes and classifications of objects in the scene, as well as the offsets of the subsequent layer. This process continues iteratively. In our work, we employ two layers of \textit{PASS}. 
Compared to one layer of \textit{PASS}, two layers of iteration mitigate the potential uncertainty in offsets that may misguide the original features. 
This multi-layer approach facilitates more accurate sampling and enables us to gather comprehensive 3D spatial information, thereby improving detection performance.

Using the adaptive sampling strategy in \textit{PASS}, we leverage sampled features from the previous layer to predict the sampling points of the subsequent layer. 
This iterative process aims to supplement the information missed at the original sampling points and allows the entire feature volume to efficiently capture spatially relevant information, even in scenarios with low sampling resolution.
In our approach, we utilize two layers of \textit{PASS} to obtain the final feature volume $V_{s}^2$. 
This feature volume is then passed into our detection head, which generates the final predictions for classification, location, and centerness through convolution layers. 



\subsection{Depth-Guided Simplified Multi-Head Attention Fusion}

\begin{figure}
    \centering
    \includegraphics[width=1\linewidth]{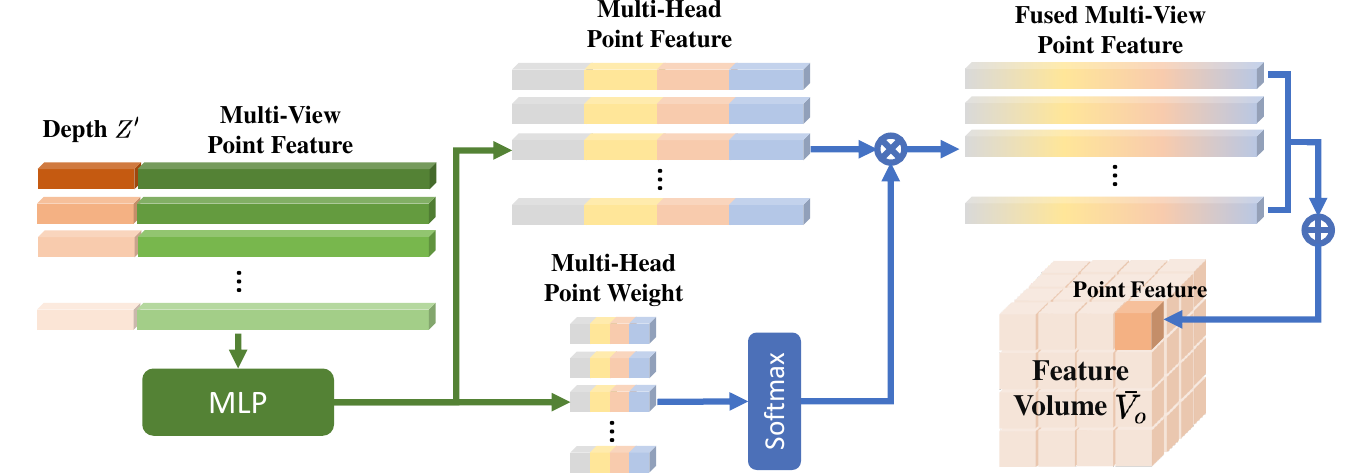}
    \caption{\textit{\textbf{Depth-Guided Simplified Multi-Head Attention Fusion}.} The process for each point in the volume is shown above. We concat the feature with the depth to predict the multi-head feature and weight. }
    \label{fig:Multi-head}
\end{figure}

The selection of an efficient and effective multi-view visual feature fusion method is crucial for 3D perception based on 2D multi-view data.
Specifically, for each sample point $\mathbf{p} \in P_{t}$, the fusion method needs to combine multi-view features ${f_i}$ sampled from ${F_i}$, ${i=1,2,\dots,N}$, where i represents the $i$-th view, into a unified feature $\hat{f}$.
In this case, in NeRF-Det, the valid features from multi-views are directly fused using an arithmetic mean. 
\begin{equation}
    \hat{f}_{\text{Naive}} =  \frac{1}{N} \sum_1^N f_i
\end{equation}
However, this straightforward approach fuses features from different views with equal weights, neglecting the issue that certain view features may be less effective due to factors such as occlusion. This approach introduces a lot of noise, and minority salient information may not be fully reflected in the final spatial features. 
The Multi-Head Attention mechanism is an alternative operator for multi-view visual feature fusion, which can be defined as:
\begin{equation}
    \hat{f}_{\text{MHA}} = \text{Softmax}(\frac{\text{Q}(f)\text{K}(f)^T}{\sqrt{d_k}})\text{V}(f)
\end{equation}
where $d_k$ is the feature dimension, $\text{Q}(f)$, $\text{K}(f)$ and $\text{V}(f)$ have shapes of the number of views multiplied by $d_k$
This fusion applies attention cross all views, allowing for comprehensive fusion. 
However, due to the fact that weight calculation involves matrix multiplication between the queries and values of multiple views, its significant computational resource requirements render it unsuitable for scenarios involving numerous viewpoints and dense sampling.
To address this issue, we propose \textit{Depth-Guided Simplified Multi-Head Attention Fusion} (\textit{DS-MHA}), as shown in Figure \ref{fig:Multi-head}, the core formula of which can be expressed as:
\begin{equation}
    \hat{f}_{\text{DS-MHA}} = \text{Softmax}(\text{W}(f, d))\text{V}(f)
\end{equation}
where $d$ is depth from each sampling point to the camera position. This approach has two advantages: 
1. Considering the dense nature of indoor scene viewpoints, directly computing the weights significantly reduces the computational cost. Additionally, because indoor scene viewpoints have a high degree of overlap, reducing interactions between viewpoints does not result in significant performance loss. 
2. We not only capture the multi-view feature $f$ of the sampling points in the scene space but also record the depth $d$, thereby making the weight prediction process depth-aware. Higher weights can be assigned to viewpoints that are truly important for each sampling point.

Specifically, the multi-view feature volume $V_{m} \in \mathbb{R} ^ {N \times W\times H \times L \times c}$ needs to be converted into a fused scene feature volume $\bar{V}_o \in \mathbb{R} ^ {W\times H \times L \times c}$.
Additionally, by employing G-MLP from NeRF branch, we can derive geometry volume $V_G \in \mathbb{R} ^ {W\times H \times L \times 1}$, which represents the opacity of the sampling points, thereby enhancing the model's geometry awareness.
We encode depth volume $Z$ into $Z^\prime \in \mathbb{R} ^ {N \times W\times H \times L \times c^{\prime}}$ through a sine encoder. Where $ c^{\prime}$ is the encoded depth feature dimension after the sine encoder. 
By concatenating the features of the sampling points and $ Z^\prime$, we pass them into a MLP to obtain the output multi-head feature volume $V_{m}^\prime \in \mathbb{R} ^ {N \times W\times H \times L \times n \times c/n}$ and the multi-head weight $W \in \mathbb{R} ^ {N \times    W\times H \times L \times n}$. In our work, the number of multi-head weights is set to $n=8$. The Softmax function is then applied to the multi-head weight tensor along the dimension of multi-views to obtain the multi-head weights for different views. 
The $V_{m}^\prime$ is then multiplied by the multi-head weights $W^\prime$ and sum the weighted feature from different views to get the $\bar{V}_o$. 
\begin{equation}
    W^{\prime} = \text{Softmax}(W), \bar{V}_o = (\sum^{N}_{i=1}W_i^{\prime} \times V_{m_i}^\prime) \\
\end{equation}
After multiplying the geometry volume $V_G$, we can obtain $V_o =  \bar{V}_o\times V_G$. 

Through the process of \textit{DS-MHA}, we can assign higher weights to the features of useful viewpoints while reasonably utilizing GPU memory. In this way, the feature fusion process will be more focused on extracting truly effective information from the space. 
\begin{figure}
    \centering
    \includegraphics[width=1\linewidth]{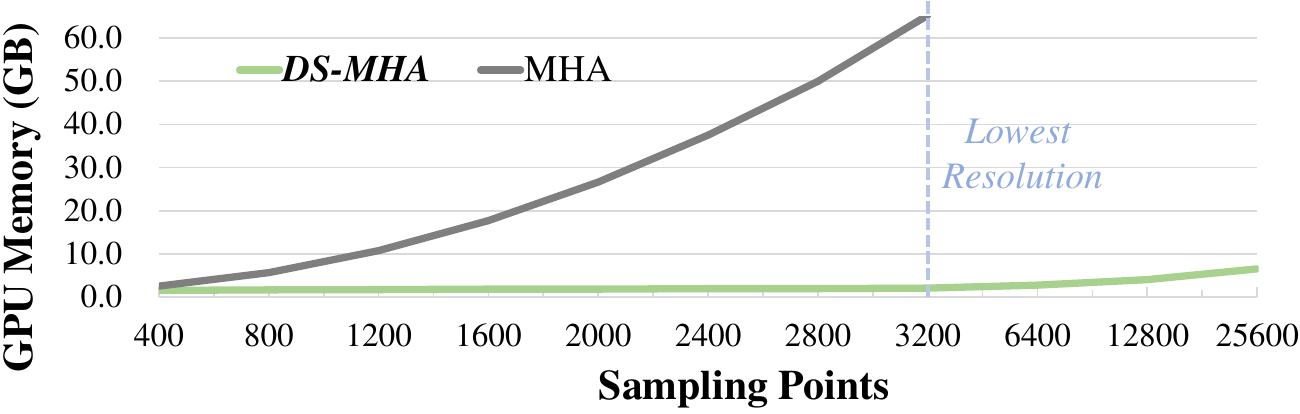}
    \caption{\textbf{Comparison of GPU Memory consumption.} Comparison between Multi-Head Attention Fusion and our \textit{Depth-Guided Simplified Multi-Head Attention Fusion}. Both evaluated under 40 views and feature dimension is 256.}
    \label{fig:GPU}
\end{figure}

\begin{table*}[!ht ]
    \centering
    \scalebox{0.68}{
        \begin{tabular}{c|cccccccccccccccccc|cc}
        \toprule

                 Methods & cab & bed & chair & sofa & table & door & wind & bkshf & pic & cntr & desk & curt & fridge & shwr & toil & sink & bath & ofurn & AP25 & AP50 \\ 
                \midrule

            Seg-Cluster   & 11.8 & 13.5 & 18.9 & 14.6 & 13.8 & 11.1 & 11.5 & 11.7 & 0.0 & 13.7 & 12.2 & 12.4 & 11.2 & 18.0 & 19.5 & 18.9 & 16.4 & 12.2 & 13.4 &  - \\ 
            Mask R-CNN    & 15.7 & 15.4 & 16.4 & 16.2 & 14.9 & 12.5 & 11.6 & 11.8 & 19.5 & 13.7 & 14.4 & 14.7 & 21.6 & 18.5 & 25.0 & 24.5 & 24.5 & 16.9 & 17.1 & 10.5 \\
            SGPN         & 20.7 & 31.5 & 31.6 & 40.6 & 31.9 & 16.6 & 15.3 & 13.6 & 0.0 & 17.4 & 14.1 & 22.2 & 0.0 & 0.0 & 72.9 & 52.4 & 0.0 & 18.6 & 22.2 &  -  \\
            3D-SIS      & 19.8 & 69.7 & 66.2 & 71.8 & 36.1 & 30.6 & 10.9 & 27.3 & 0.0 & 10.0 & 46.9 & 14.1 & 53.8 & 36.0 & 87.6 & 43.0 & 84.3 & 16.2 & 40.2 & 22.5 \\
            VoteNet      & 36.3 & 87.9 & 88.7 & 89.6 & 58.8 & 47.3 & 38.1 & 44.6 & 7.8 & 56.1 & 71.7 & 47.2 & 45.4 & 57.1 & 94.9 & 54.7 & 92.1 & 37.2 & 58.7 & 33.5 \\
            FCAF3D      & 57.2 & 87.0 & 95.0 & 92.3 & 70.3 & 61.1 & 60.2 & 64.5 & 29.9 & 64.3 & 71.5 & 60.1 & 52.4 & 83.9 & 99.9 & 84.7 & 86.6 & 65.4 & 71.5 & 57.3 \\
            CAGroup3D    & 60.4 & 93.0 & 95.3 & 92.3 & 69.9 & 67.9 & 63.6 & 67.3 & 40.7 & 77.0 & 83.9 & 69.4 & 65.7 & 73.0 & 100 & 79.7 & 87.0 & 66.1 & 75.1& 61.3 \\ \midrule
            ImVoxelNet    & 34.5 & 83.6 & 72.6 & 71.6 & 54.2 & 30.3 & 14.8 & 42.6 & 0.8  & 40.8 & 65.3 & 18.3 & 52.2 & 40.9 & 90.4 & 53.3 & 74.9 & 33.1 & 48.4 & 23.7 \\
            NeRF-Det*    & 37.7 & \textbf{84.1}& 74.5 & 71.8 & 54.2 & 34.2 & 17.4 & 51.6 & 0.1  & 54.2 & 71.3 & 16.7 & 54.5 & 55.0 & 92.1 & 50.7 & 73.8 & 34.1 & 51.8 & 27.4 \\
            NeRF-Det++*   & 38.9 & 83.9 & 73.9 & 77.6 & \textbf{57.2}& 33.3 & 23.5 & 47.9 & 1.5  & 56.9 & \textbf{77.7}& 21.1 & \textbf{61.5}& 46.8 & 92.8 & 49.2 & 80.2 & 34.5 & 53.2 & 29.6 \\
    \rowcolor{lgray}  \nds{}*(Ours)   & \textbf{43.7}& 83.8 & \textbf{78.3}& \textbf{83.0}& 56.8 & \textbf{43.2}& \textbf{28.8}& \textbf{52.0}& \textbf{3.8 }& \textbf{70.5}& 69.8 & \textbf{28.5}& 59.4 & \textbf{61.6}& \textbf{93.1}& \textbf{52.4}& \textbf{83.3}& \textbf{44.5}& \textbf{57.6}& \textbf{35.6}\\
    \bottomrule 
        \end{tabular}
}
    \caption{\textbf{Quantitative comparison} on ScanNetV2 with Seg-Cluster \cite{Sgpn}, Mask R-CNN \cite{Maskrcnn}, SGPN \cite{Sgpn}, 3D-SIS \cite{3dsis}, VoteNet \cite{Votenet}, FCAF3D \cite{FCAF3D}, CAGroup3D \cite{cagroup3d}, ImVoxelNet \cite{imvoxelnet}, NeRF-Det \cite{NeRF-Det} and NeRF-Det++ \cite{NeRF-Detpp}. The above block shows the methods that base on the point cloud or RGBD input. The block below shows methods base on the multi-view RGB images input. \text{*} indicates the method with depth supervision. For the methods base on the multi-view RGB images, we all evaluate on the result of ResNet50. We assess performance of each category using AP25 as the evaluation metric. We use bold to highlight the method in best performance of each category.}
    \label{tab:comparison}
\end{table*}

\subsection{Training Objective}
In the detection branch, we supervise the output of classification, location, and centerness in a manner similar to ImVoxelNet \cite{imvoxelnet}, using FCOS3D-structured head. 
The output of each layer of \textit{PASS} is supervised.
This ensures that each layer comprehends spatial information while aiding in the improvement of offset predictions for subsequent layers. In NeRF branch, depth and RGB are supervised to capture spatial information.
Therefore, in our training process, we use a total of three layers of losses $L_{cls}$ $, L_{cntr },L_{loc }$ for classification loss, centerness loss, and localization loss, to supervise the detection process. In the NeRF branch, we use RGB and depth losses $L_{c}$ and $L_{d}$ as rendering loss, $L_{c}=\left\|\hat{C}(r)-\hat{C}_{g t}(r)\right\|_{2}$, $L_{d}=\left\|D(r)-D_{g t}(r)\right\|$. NeRF branch only needs to render RGB during training, and during inference, it only requires the  G-MLP to output the density of sampling points in space. 
\begin{equation}
    L=\sum_{i=1}^{3}(L_{\text{cls}_i}+L_{cntr_i}+L_{loc_i})+L_{c}+L_{d}
\end{equation}

\section{Experiments}

\subsection{Experimental Setting}

\noindent \textbf{Dataset.} 
In our experiment, we closely adhere to the complete structure of NeRF-Det \cite{NeRF-Det}, with the goal of validating the effectiveness of \nds{} in enhancing high-quality object detection within the NeRF branch framework, while minimizing alterations to the original network architecture. We conduct experiments using the ScannetV2 \cite{scannet}, which is a widely utilized benchmark in indoor object detection and other indoor perception tasks. It contains 1513 indoor scenes of various kinds of indoor rooms. Following the official separation, we use 1201 scenes for training and 312 scenes for validation and testing. Utilizing tools provided by MMDetection3D, we randomly select 300 RGB images and their corresponding depth images at a resolution of $320\times240$ for each scene. It is important to note that the depth information is used exclusively during the training phase to supervise the output of the NeRF branch. 
To make our results more convincing, we also conducted experiments on ARKITScenes \cite{dehghan2021arkitscenes}. Since the orientations of the ARKITScenes are diverse, we selected a total of 2414 scenes for training and 309 scenes for testing, specifically from the Top and Down directions. The resolution of the input images is $256\times192$.

\noindent \textbf{Implement Details.} 
We assess the performance of the methods using mean Average Precision (mAP) and mean Average Recall (mAR) at 0.25, 0.5, and 0.7 Intersection over Union (IoU) thresholds, denoted as AP25 / AP50 / AP70 and AR25 / AR50 / AR70, respectively. IoU70 is specifically used as a metric in PARQ \cite{PARQ}. For each scene, we randomly select 50 images for training and 101 images for testing from the processed set of 300 images. Given the simplicity of the original network architecture, the training process converges quickly. To ensure robust performance, we extend the number of training epochs by an additional 50\%, preventing overfitting while allowing our methods to converge effectively. This strategy facilitates a more accurate evaluation of the network and yields more reliable results.

\subsection{Main Results}

We use ResNet50 as the camera encoder (i.e., backbone) for our experiment. A comprehensive quantitative comparison is presented in Table \ref{tab:comparison}. 
We use NeRF-Det \cite{NeRF-Det} with ResNet50 and depth supervision as our baseline, employing the same training strategy for comparison, as shown in Table \ref{tab:compare}. 
Under identical training conditions, our full method significantly improves performance, with a mean Average Precision (mAP) increase of +5.02\% under IoU25 and +5.92\% under IoU50 compared to the baseline. 
These results indicate that our method can effectively enhance object detection performance. The notable improvement under IoU50 demonstrates our method's ability to better integrate spatial information from multi-view perspectives and adaptively sample to supplement the original sampling points. By adopting an adaptive sampling approach, we have substantially improved the effectiveness of the sampling process.
Our method outperforms previous fixed sampling approaches, demonstrating its ability to effectively supplement missing information in the original pipeline and improve the quality of predicting bounding boxes.

\begin{table}
    \centering
    \scalebox{0.7}{
    \begin{tabular}{l |c|c|c|c|c|c}
    \toprule
 Method & {AP25} & {AR25}& {AP50}& {AR50}& {AP70}& {AR70}\\ 
\midrule

NeRF-Det           & 52.6&  67.3&  29.7&  40.8 &09.8 &16.1 \\
\nds{}&\textbf{57.6}  &\textbf{71.4}&\textbf{35.6}&\textbf{47.0} &\textbf{13.5} &\textbf{21.4} \\
\bottomrule

    \end{tabular}
    }
    \caption{Comparison between NeRF-Det and \nds{} under the same training strategy on ScanNet.}
    \label{tab:compare}
\end{table}

The code for the ARKITScenes part is not open-sourced in ScanNet, so we referred to the setup of ImGeoNet \cite{tu2023imgeonet} and used the top and down direction scenes from ARKITScenes. The model setup here is identical to that in the original NeRF-Det paper. As can be seen in Table \ref{tab:ARKITScenes} compared to the NeRF-Det method, our approach also demonstrates consistent improvements on ARKITScenes.

\begin{table}
    \centering
    \scalebox{0.7}{
    \begin{tabular}{l |c|c|c|c}
    \toprule
    Method & {AP25} & {AR25} & {AP50} & {AR50} \\ 
    \midrule

    NeRF-Det           & 52.1 & 64.2 & 21.1 & 32.4 \\
    \nds{}                & \textbf{53.9}  & \textbf{67.8} & \textbf{23.9}  & \textbf{35.4}  \\
    \bottomrule

    \end{tabular}
    }
    \caption{Comparison between NeRF-Det and our method on ARKITScenes.}
    \label{tab:ARKITScenes}
\end{table}

We quantified the number of sampled points falling within the 3D GT boxes in \textit{PASS} layer on the ScanNetV2 dataset, as shown in Table \ref{tab:scene_comparison}. This effectively demonstrates the effectiveness of our sampling strategy. 

\begin{table}[t]
\addtolength{\tabcolsep}{-2pt}

    \centering
    \scalebox{0.6}{
    \begin{tabular}{lcccc|c}
    \toprule

        Method   & scene0131\_00 & scene0575\_01 & scene0050\_02 & scene0357\_01 & Means of All Scenes \\ 
    \midrule

        NeRF-Det & 3902 & 2093 & 1984 & 856 & 1611.5 \\
        Ours & 5024\textcolor{Green}{\,+28.8\%} & 2931\textcolor{Green}{\,+40.0\%} & 2874\textcolor{Green}{\,+44.9\%} & 1299\textcolor{Green}{\,+51.8\%} & 1992.3\textcolor{Green}{\,+23.6\%} \\
    \bottomrule

    \end{tabular}}
    \caption{Number of sampling points fall within the GT 3D bounding box for NeRF-Det and our method.}
    \label{tab:scene_comparison}
\end{table}

\subsection{Ablation Study}
\noindent \textbf{Key Components Ablation.} 
In our experiments, we introduce flip augmentation by horizontally flipping the posed images and their corresponding depths. Additionally, we mirror-reflect the camera extrinsics along the x-y plane. This approach not only allows us to mirror-flip the 2D images but also to mirror-transform the sampled 3D voxels. Throughout the entire pipeline, whether in 2D or 3D networks, this method facilitates convenient augmentation, thereby enhancing the diversity of the network's input and improving its robustness.
In Table \ref{tab:ablation}, we present the ablation on key components in \nds{}. 
We incrementally introduce various components, including \textit{DS-MHA}, Flip Aug, 1 Layer \textit{PASS}, and 2 Layers \textit{PASS}. 
It is evident that with the introduction of components, there is a significant enhancement in detection performance. Compared to directly averaging the features at a position, \textit{DS-MHA}  greatly enhanced the detection process. Flip Aug and +1 Layer \textit{PASS} configurations also show further boosts, although +1 Layer \textit{PASS} shows a slight decrease in AR25 and AR50.
For the +2 Layers \textit{PASS} configuration, the best performance is observed across all IoU thresholds. This indicates that introducing two additional layers of adaptive sampling contributes to improving the robustness and accuracy of the detection model in \nds{}. Specifically, the adoption of a 2-layer \textit{PASS} helps address the uncertainty that may arise from a 1-layer \textit{PASS}. This uncertainty could result in ineffective or incorrect features, impacting the final detection performance by sampling positions irrelevant to the object itself.

\noindent \textbf{Resolution Study.} 
Table \ref{tab:res} presents the performance evaluation of our method across various resolutions, aimed at assessing its robustness and efficiency under different spatial sampling configurations. 
When using the same resolution, the total number of sampling points is tripled compared to NeRF-Det \cite{NeRF-Det} with a 2-layer \textit{PASS}. These variations allow us to analyze how changes in resolution influence our method's performance.
\begin{table}
    \centering
        \scalebox{0.7}{
    \begin{tabular}{ccc|cc|cc} 
    \toprule



\textit{DS-MHA}&Flip Aug& \textit{PASS}& AP25& AR25& AP50& AR50\\
\midrule

          &            &           & 52.6& 67.3& 29.7& 40.8 \\ 
\checkmark&            &           & 55.7& 70.8& 32.1& 43.8 \\
\checkmark& \checkmark &           & 56.7& \textbf{73.1}& 34.5& 46.2 \\
\checkmark& \checkmark &  +1       & 57.4& 70.9& 34.7& 45.9 \\
\checkmark& \checkmark &  +2       & \textbf{57.6}& 71.4& \textbf{35.6}& \textbf{47.0} \\

\bottomrule

    \end{tabular}}
    \caption{\textbf{Ablation on key components. }Ablation on the \textit{DS-MHA}, flip augmentation, and the layer number of \textit{PASS}.}
    \label{tab:ablation}
\end{table}

\begin{table}
\addtolength{\tabcolsep}{-2pt}

    \centering
    \scalebox{0.65}{
    \begin{tabular}{l|cc|cc|cc}
    \toprule

        Method   & AP25  & AR25   & AP50   & AR50   & AP70   & AR70   \\ 
\midrule

        NeRF-Det& 52.6 & 67.3 & 29.7 & 40.8 & 09.8 & 16.1\\
        Ours& 57.6\textcolor{Green}{\,+5.0}& 71.4\textcolor{Green}{\,+4.1} & 35.6\textcolor{Green}{\,+5.9}& 47.0\textcolor{Green}{\,+6.2}& 13.5\textcolor{Green}{\,+3.7}& 21.4\textcolor{Green}{\,+4.3}\\\midrule
        
        NeRF-Det /2& 49.7 & 65.9 & 26.1 & 38.3 & 07.8 & 14.0\\
        Ours /2& 54.3\textcolor{Green}{\,+4.6} & 68.4\textcolor{Green}{\,+2.5}& 28.9\textcolor{Green}{\,+2.8}& 41.4\textcolor{Green}{\,+3.1}& 10.6\textcolor{Green}{\,+2.8}& 18.0\textcolor{Green}{\,+4.0}  \\\midrule
        
        NeRF-Det /4& 46.8& 61.6& 25.2& 34.2& 07.9& 13.8\\ 
        Ours /4& 51.6\textcolor{Green}{\,+4.8}& 66.2\textcolor{Green}{\,+5.4}& 29.6\textcolor{Green}{\,+4.4}& 40.5\textcolor{Green}{\,+6.3}& 10.0\textcolor{Green}{\,+2.1}& 16.6\textcolor{Green}{\,+2.8}\\\midrule
        
        NeRF-Det /8& 44.7& 58.3& 23.9& 32.8& 07.8& 13.0\\
        Ours /8& 50.9\textcolor{Green}{\,+6.1} & 63.4\textcolor{Green}{\,+5.1}  & 29.1\textcolor{Green}{\,+5.2}  & 39.0\textcolor{Green}{\,+6.2}  & 10.8\textcolor{Green}{\,+3.0}  & 17.2\textcolor{Green}{\,+4.2}\\
        
\bottomrule

    \end{tabular}}
    \caption{\textbf{Resolution study.} Comparison of performance with NeRF-Det and \nds{} across different resolutions.}
    \label{tab:res}
\end{table}

In our experiment, we employ \textit{PASS} to capture features in the spatial domain. Our method excels at enhancing the sampling of effective positional information, particularly in low-resolution scenarios.
By utilizing only 3/8 or 3/4 of original NeRF-Det's sampling points, we attain performance that match or exceed those of NeRF-Det at IoU50 threshold. Interestingly, even when reducing the number of sampling points, our method yields similar results, suggesting that our approach maintains robustness in low spatial resolutions without the need to increase the resolution indiscriminately.
Our experiments reveal a positive correlation between mAR and the number of sampling points for indoor scenes. Even with fewer sampling points than NeRF-Det, our method maintains a higher recall rate.
Furthermore, at an IoU70 threshold, our method outperforms NeRF-Det even with the lowest resolution sampling points. This capability enables our model to effectively distinguish objects in challenging scenarios, such as overlapping or partially occluded targets, through progressive adaptive sampling. Consequently, our approach maintains outstanding detection accuracy in complex spatial environments.


\begin{table}
    \centering
            \scalebox{0.7}{

    \begin{tabular}{l|cc|cc|cc} 
        \toprule

Method & AP25& AR25& AP50& AR50& AP70& AR70 \\ 
\midrule

w/o offset& 51.6  & 66.5  & 28.4  & 39.9  & 11.0 & 17.9  \\ 
w/ \; offset& 57.6  & 71.4  & 35.6  & 47.0  & 13.5 &  21.4  \\
\bottomrule

    \end{tabular}}
    \caption{\textbf{Offset Ablation}. In without offset setting, we sample at the original sampling points for three times.}
    \label{tab:offset}
\end{table}

\noindent \textbf{Offset Ablation. }
We aim to retain all network components but abandon the spatial sampling process utilizing offsets, shown in Table \ref{tab:offset}. This implies that we sample at points for three times at the same positions. 
Despite an increase in the number of parameters, we observed that performance not only fails to improve but actually deteriorates due to overfitting on the training data. This confirms that the offset strategy of \textit{PASS} indeed captures useful information from other spatial positions to supplement the features of the original sampling positions and helps prevent network overfitting.

\begin{table}

    \centering
    \scalebox{0.7}{
        \begin{tabular}{cc|cc|cc} 
            \toprule

     \multicolumn{2}{c|}{Method}& \multicolumn{2}{c|}{\nds{}} & \multicolumn{2}{c}{\nds{}  /8} \\
        w/ $Z$ & head & AP50& AR50 & AP50& AR50 \\
        \midrule
        $\checkmark$ & 4         & 34.4\textcolor{Red}{\,-1.2}  & 46.4\textcolor{Red}{\,-0.6} & 28.7\textcolor{Red}{\,-0.4} & 38.6\textcolor{Red}{\,-0.4}      \\ 
        $\checkmark$ & 16        & 32.3\textcolor{Red}{\,-3.3}  & 44.0\textcolor{Red}{\,-3.0} & 28.2\textcolor{Red}{\,-0.9} & 37.7\textcolor{Red}{\,-1.3}      \\
        $\times$ & 8             & 31.9\textcolor{Red}{\,-3.7}  & 43.2\textcolor{Red}{\,-3.8} & 28.5\textcolor{Red}{\,-0.6} & 37.9\textcolor{Red}{\,-1.1}      \\
        \midrule
        $\checkmark$ & 8   & \textbf{35.6}  & \textbf{47.0} & \textbf{29.1} & \textbf{39.0}      \\
        \bottomrule
        
    \end{tabular}
    }
    \caption{\textbf{Ablation on the number of heads and the presence of  Depth} $Z$ . We conduct experiments at the original resolution and at 1/8 resolution of \nds{}. }
    \label{tab: head&z}
\end{table}

\noindent \textbf{Head \& Depth Ablation.}
We present the experimental results under different resolutions with varying numbers of heads and the presence or absence of distance weight in Table \ref{tab: head&z}. It can be seen that the model performs best when the number of heads is 8 and distance weight is applied. Additionally, we found that the number of heads and the presence of distance weighting have a greater impact on performance at higher resolutions. We believe this is because, at higher resolutions, the perception process is more sensitive to noise, leading to significant effects from minor disturbances on the results.This also demonstrates the effectiveness of \textit{DS-MHA} in reducing noise when fusing features from different views.

\section{Conclusion}
This paper introduces the \nds{} to enhance the performance of object detection for multi-view images in a continuous NeRF-based representation. To sufficiently leverage the advantages brought by the NeRF branch to the perceptual process, we employ the \textit{Progressive Adaptive Sampling Strategy}, which fully utilize the notable characteristic of continuity in NeRF-based representation. Additionally, recognizing limitations during the fusion process of multi-view features in space, we propose the \textit{Depth-Guided Simplified Multi-Head Attention Fusion}. This approach utilizes weights to address situations where a particular viewpoint in space may be occluded in the presence of multiple perspectives. Extensive experiments on the ScanNetV2 and ARKITScenes dataset validate the effectiveness of our method in improving the performance of detection tasks.

{\small
\bibliography{aaai25.bib}

\begin{thebibliography}{45}
\providecommand{\natexlab}[1]{#1}

\bibitem[{Barron et~al.(2021)Barron, Mildenhall, Tancik, Hedman, Martin-Brualla, and Srinivasan}]{mipnerf}
Barron, J.~T.; Mildenhall, B.; Tancik, M.; Hedman, P.; Martin-Brualla, R.; and Srinivasan, P.~P. 2021.
\newblock Mip-nerf: A multiscale representation for anti-aliasing neural radiance fields.
\newblock In \emph{Proceedings of the IEEE/CVF International Conference on Computer Vision}, 5855--5864.

\bibitem[{Baruch et~al.(2021)Baruch, Chen, Dehghan, Dimry, Feigin, Fu, Gebauer, Joffe, Kurz, Schwartz, and Shulman}]{dehghan2021arkitscenes}
Baruch, G.; Chen, Z.; Dehghan, A.; Dimry, T.; Feigin, Y.; Fu, P.; Gebauer, T.; Joffe, B.; Kurz, D.; Schwartz, A.; and Shulman, E. 2021.
\newblock {ARK}itScenes - A Diverse Real-World Dataset for 3D Indoor Scene Understanding Using Mobile {RGB}-D Data.
\newblock In \emph{Thirty-fifth Conference on Neural Information Processing Systems Datasets and Benchmarks Track (Round 1)}.

\bibitem[{Carion et~al.(2020)Carion, Massa, Synnaeve, Usunier, Kirillov, and Zagoruyko}]{DETR}
Carion, N.; Massa, F.; Synnaeve, G.; Usunier, N.; Kirillov, A.; and Zagoruyko, S. 2020.
\newblock End-to-end object detection with transformers.
\newblock In \emph{European conference on computer vision}, 213--229. Springer.

\bibitem[{Chen et~al.(2021)Chen, Xu, Zhao, Zhang, Xiang, Yu, and Su}]{mvsnerf}
Chen, A.; Xu, Z.; Zhao, F.; Zhang, X.; Xiang, F.; Yu, J.; and Su, H. 2021.
\newblock Mvsnerf: Fast generalizable radiance field reconstruction from multi-view stereo.
\newblock In \emph{Proceedings of the IEEE/CVF International Conference on Computer Vision}, 14124--14133.

\bibitem[{Dai et~al.(2017)Dai, Chang, Savva, Halber, Funkhouser, and Nie{\ss}ner}]{scannet}
Dai, A.; Chang, A.~X.; Savva, M.; Halber, M.; Funkhouser, T.; and Nie{\ss}ner, M. 2017.
\newblock Scannet: Richly-annotated 3d reconstructions of indoor scenes.
\newblock In \emph{Proceedings of the IEEE conference on computer vision and pattern recognition}, 5828--5839.

\bibitem[{Ding, Han, and Niethammer(2019)}]{Votenet}
Ding, Z.; Han, X.; and Niethammer, M. 2019.
\newblock Votenet: A deep learning label fusion method for multi-atlas segmentation.
\newblock In \emph{Medical Image Computing and Computer Assisted Intervention--MICCAI 2019: 22nd International Conference, Shenzhen, China, October 13--17, 2019, Proceedings, Part III 22}, 202--210. Springer.

\bibitem[{Fridovich-Keil et~al.(2022)Fridovich-Keil, Yu, Tancik, Chen, Recht, and Kanazawa}]{plenoxels}
Fridovich-Keil, S.; Yu, A.; Tancik, M.; Chen, Q.; Recht, B.; and Kanazawa, A. 2022.
\newblock Plenoxels: Radiance fields without neural networks.
\newblock In \emph{Proceedings of the IEEE/CVF Conference on Computer Vision and Pattern Recognition}, 5501--5510.

\bibitem[{Fu et~al.(2022)Fu, Zhang, Chen, Lu, Zhu, Zhou, Geiger, and Liao}]{panopticnerf}
Fu, X.; Zhang, S.; Chen, T.; Lu, Y.; Zhu, L.; Zhou, X.; Geiger, A.; and Liao, Y. 2022.
\newblock Panoptic nerf: 3d-to-2d label transfer for panoptic urban scene segmentation.
\newblock In \emph{2022 International Conference on 3D Vision (3DV)}, 1--11. IEEE.

\bibitem[{He et~al.(2017)He, Gkioxari, Doll{\'a}r, and Girshick}]{Maskrcnn}
He, K.; Gkioxari, G.; Doll{\'a}r, P.; and Girshick, R. 2017.
\newblock Mask r-cnn.
\newblock In \emph{Proceedings of the IEEE international conference on computer vision}, 2961--2969.

\bibitem[{Hou, Dai, and Nie{\ss}ner(2019)}]{3dsis}
Hou, J.; Dai, A.; and Nie{\ss}ner, M. 2019.
\newblock 3d-sis: 3d semantic instance segmentation of rgb-d scans.
\newblock In \emph{Proceedings of the IEEE/CVF conference on computer vision and pattern recognition}, 4421--4430.

\bibitem[{Hu et~al.(2023)Hu, Huang, Liu, Tai, and Tang}]{NeRF-RPN}
Hu, B.; Huang, J.; Liu, Y.; Tai, Y.-W.; and Tang, C.-K. 2023.
\newblock NeRF-RPN: A general framework for object detection in NeRFs.
\newblock In \emph{Proceedings of the IEEE/CVF Conference on Computer Vision and Pattern Recognition}, 23528--23538.

\bibitem[{Huang et~al.(2024{\natexlab{a}})Huang, Hou, Ye, Huang, Huang, Lin, Cai, and Ouyang}]{NeRF-Detpp}
Huang, C.; Hou, Y.; Ye, W.; Huang, D.; Huang, X.; Lin, B.; Cai, D.; and Ouyang, W. 2024{\natexlab{a}}.
\newblock NeRF-Det++: Incorporating Semantic Cues and Perspective-aware Depth Supervision for Indoor Multi-View 3D Detection.
\newblock \emph{arXiv preprint arXiv:2402.14464}.

\bibitem[{Huang et~al.(2021)Huang, Huang, Zhu, Ye, and Du}]{bevdet}
Huang, J.; Huang, G.; Zhu, Z.; Ye, Y.; and Du, D. 2021.
\newblock Bevdet: High-performance multi-camera 3d object detection in bird-eye-view.
\newblock \emph{arXiv preprint arXiv:2112.11790}.

\bibitem[{Huang et~al.(2024{\natexlab{b}})Huang, Li, Sima, Wang, Wang, Qiao, and Li}]{huang2024leveraging}
Huang, L.; Li, Z.; Sima, C.; Wang, W.; Wang, J.; Qiao, Y.; and Li, H. 2024{\natexlab{b}}.
\newblock Leveraging vision-centric multi-modal expertise for 3d object detection.
\newblock \emph{Advances in Neural Information Processing Systems}, 36.

\bibitem[{Huang et~al.(2023{\natexlab{a}})Huang, Wang, Zeng, Zhang, Cao, Ji, Yan, and Li}]{huang2023geometric}
Huang, L.; Wang, H.; Zeng, J.; Zhang, S.; Cao, L.; Ji, R.; Yan, J.; and Li, H. 2023{\natexlab{a}}.
\newblock Geometric-aware Pretraining for Vision-centric 3D Object Detection.
\newblock \emph{arXiv preprint arXiv:2304.03105}.

\bibitem[{Huang et~al.(2023{\natexlab{b}})Huang, Zheng, Zhang, Zhou, and Lu}]{tpvformer}
Huang, Y.; Zheng, W.; Zhang, Y.; Zhou, J.; and Lu, J. 2023{\natexlab{b}}.
\newblock Tri-perspective view for vision-based 3d semantic occupancy prediction.
\newblock In \emph{Proceedings of the IEEE/CVF Conference on Computer Vision and Pattern Recognition}, 9223--9232.

\bibitem[{Irshad et~al.(2024)Irshad, Zakharov, Guizilini, Gaidon, Kira, and Ambrus}]{nerfmae}
Irshad, M.~Z.; Zakharov, S.; Guizilini, V.; Gaidon, A.; Kira, Z.; and Ambrus, R. 2024.
\newblock NeRF-MAE: Masked AutoEncoders for Self-Supervised 3D Representation Learning for Neural Radiance Fields.
\newblock In \emph{European Conference on Computer Vision (ECCV)}.

\bibitem[{Jeong et~al.(2022)Jeong, Shin, Lee, Choy, Anandkumar, Cho, and Park}]{PeRFception}
Jeong, Y.; Shin, S.; Lee, J.; Choy, C.; Anandkumar, A.; Cho, M.; and Park, J. 2022.
\newblock PeRFception: Perception using radiance fields.
\newblock \emph{Advances in Neural Information Processing Systems}, 35: 26105--26121.

\bibitem[{Kerbl et~al.(2023)Kerbl, Kopanas, Leimk{\"u}hler, and Drettakis}]{Gaussian}
Kerbl, B.; Kopanas, G.; Leimk{\"u}hler, T.; and Drettakis, G. 2023.
\newblock 3D Gaussian Splatting for Real-Time Radiance Field Rendering.
\newblock \emph{ACM Transactions on Graphics}, 42(4).

\bibitem[{Li et~al.(2023)Li, Yu, Choy, Xiao, Alvarez, Fidler, Feng, and Anandkumar}]{voxformer}
Li, Y.; Yu, Z.; Choy, C.; Xiao, C.; Alvarez, J.~M.; Fidler, S.; Feng, C.; and Anandkumar, A. 2023.
\newblock Voxformer: Sparse voxel transformer for camera-based 3d semantic scene completion.
\newblock In \emph{Proceedings of the IEEE/CVF Conference on Computer Vision and Pattern Recognition}, 9087--9098.

\bibitem[{Li et~al.(2022)Li, Wang, Li, Xie, Sima, Lu, Qiao, and Dai}]{bevformer}
Li, Z.; Wang, W.; Li, H.; Xie, E.; Sima, C.; Lu, T.; Qiao, Y.; and Dai, J. 2022.
\newblock Bevformer: Learning bird’s-eye-view representation from multi-camera images via spatiotemporal transformers.
\newblock In \emph{European conference on computer vision}, 1--18. Springer.

\bibitem[{Liu et~al.(2022{\natexlab{a}})Liu, Lu, Xu, Liu, Li, and Chen}]{camliflow}
Liu, H.; Lu, T.; Xu, Y.; Liu, J.; Li, W.; and Chen, L. 2022{\natexlab{a}}.
\newblock CamLiFlow: Bidirectional camera-LiDAR fusion for joint optical flow and scene flow estimation.
\newblock In \emph{Proceedings of the IEEE/CVF Conference on Computer Vision and Pattern Recognition}, 5791--5801.

\bibitem[{Liu et~al.(2023{\natexlab{a}})Liu, Teng, Lu, Wang, and Wang}]{sparsebev}
Liu, H.; Teng, Y.; Lu, T.; Wang, H.; and Wang, L. 2023{\natexlab{a}}.
\newblock Sparsebev: High-performance sparse 3d object detection from multi-camera videos.
\newblock In \emph{Proceedings of the IEEE/CVF International Conference on Computer Vision}, 18580--18590.

\bibitem[{Liu et~al.(2020)Liu, Gu, Zaw~Lin, Chua, and Theobalt}]{nsvf}
Liu, L.; Gu, J.; Zaw~Lin, K.; Chua, T.-S.; and Theobalt, C. 2020.
\newblock Neural sparse voxel fields.
\newblock \emph{Advances in Neural Information Processing Systems}, 33: 15651--15663.

\bibitem[{Liu et~al.(2022{\natexlab{b}})Liu, Wang, Zhang, and Sun}]{petr}
Liu, Y.; Wang, T.; Zhang, X.; and Sun, J. 2022{\natexlab{b}}.
\newblock Petr: Position embedding transformation for multi-view 3d object detection.
\newblock In \emph{European Conference on Computer Vision}, 531--548. Springer.

\bibitem[{Liu et~al.(2023{\natexlab{b}})Liu, Tang, Amini, Yang, Mao, Rus, and Han}]{bevfusion}
Liu, Z.; Tang, H.; Amini, A.; Yang, X.; Mao, H.; Rus, D.~L.; and Han, S. 2023{\natexlab{b}}.
\newblock Bevfusion: Multi-task multi-sensor fusion with unified bird's-eye view representation.
\newblock In \emph{2023 IEEE international conference on robotics and automation (ICRA)}, 2774--2781. IEEE.

\bibitem[{Mildenhall et~al.(2021)Mildenhall, Srinivasan, Tancik, Barron, Ramamoorthi, and Ng}]{nerf}
Mildenhall, B.; Srinivasan, P.~P.; Tancik, M.; Barron, J.~T.; Ramamoorthi, R.; and Ng, R. 2021.
\newblock Nerf: Representing scenes as neural radiance fields for view synthesis.
\newblock \emph{Communications of the ACM}, 65(1): 99--106.

\bibitem[{M{\"u}ller et~al.(2022)M{\"u}ller, Evans, Schied, and Keller}]{INGP}
M{\"u}ller, T.; Evans, A.; Schied, C.; and Keller, A. 2022.
\newblock Instant neural graphics primitives with a multiresolution hash encoding.
\newblock \emph{ACM Transactions on Graphics (ToG)}, 41(4): 1--15.

\bibitem[{Niemeyer et~al.(2022)Niemeyer, Barron, Mildenhall, Sajjadi, Geiger, and Radwan}]{regnerf}
Niemeyer, M.; Barron, J.~T.; Mildenhall, B.; Sajjadi, M.~S.; Geiger, A.; and Radwan, N. 2022.
\newblock Regnerf: Regularizing neural radiance fields for view synthesis from sparse inputs.
\newblock In \emph{Proceedings of the IEEE/CVF Conference on Computer Vision and Pattern Recognition}, 5480--5490.

\bibitem[{Qi et~al.(2017)Qi, Su, Mo, and Guibas}]{pointnet}
Qi, C.~R.; Su, H.; Mo, K.; and Guibas, L.~J. 2017.
\newblock Pointnet: Deep learning on point sets for 3d classification and segmentation.
\newblock In \emph{Proceedings of the IEEE conference on computer vision and pattern recognition}, 652--660.

\bibitem[{Rukhovich, Vorontsova, and Konushin(2022{\natexlab{a}})}]{FCAF3D}
Rukhovich, D.; Vorontsova, A.; and Konushin, A. 2022{\natexlab{a}}.
\newblock FCAF3D: Fully Convolutional Anchor-Free 3D Object Detection.
\newblock In \emph{European Conference on Computer Vision}, 477--493.

\bibitem[{Rukhovich, Vorontsova, and Konushin(2022{\natexlab{b}})}]{imvoxelnet}
Rukhovich, D.; Vorontsova, A.; and Konushin, A. 2022{\natexlab{b}}.
\newblock Imvoxelnet: Image to voxels projection for monocular and multi-view general-purpose 3d object detection.
\newblock In \emph{Proceedings of the IEEE/CVF Winter Conference on Applications of Computer Vision}, 2397--2406.

\bibitem[{Sun, Sun, and Chen(2022)}]{dvgo}
Sun, C.; Sun, M.; and Chen, H.-T. 2022.
\newblock Direct voxel grid optimization: Super-fast convergence for radiance fields reconstruction.
\newblock In \emph{Proceedings of the IEEE/CVF Conference on Computer Vision and Pattern Recognition}, 5459--5469.

\bibitem[{Tu et~al.(2023)Tu, Chuang, Liu, Sun, Zhang, Roy, Kuo, and Sun}]{tu2023imgeonet}
Tu, T.; Chuang, S.-P.; Liu, Y.-L.; Sun, C.; Zhang, K.; Roy, D.; Kuo, C.-H.; and Sun, M. 2023.
\newblock Imgeonet: Image-induced geometry-aware voxel representation for multi-view 3d object detection.
\newblock In \emph{Proceedings of the IEEE/CVF International Conference on Computer Vision}, 6996--7007.

\bibitem[{Wang et~al.(2022{\natexlab{a}})Wang, Dong, Shi, Li, Li, Li, Wang et~al.}]{cagroup3d}
Wang, H.; Dong, S.; Shi, S.; Li, A.; Li, J.; Li, Z.; Wang, L.; et~al. 2022{\natexlab{a}}.
\newblock Cagroup3d: Class-aware grouping for 3d object detection on point clouds.
\newblock \emph{Advances in Neural Information Processing Systems}, 35: 29975--29988.

\bibitem[{Wang et~al.(2021{\natexlab{a}})Wang, Liu, Liu, Theobalt, Komura, and Wang}]{neus}
Wang, P.; Liu, L.; Liu, Y.; Theobalt, C.; Komura, T.; and Wang, W. 2021{\natexlab{a}}.
\newblock Neus: Learning neural implicit surfaces by volume rendering for multi-view reconstruction.
\newblock \emph{arXiv preprint arXiv:2106.10689}.

\bibitem[{Wang et~al.(2021{\natexlab{b}})Wang, Wang, Genova, Srinivasan, Zhou, Barron, Martin-Brualla, Snavely, and Funkhouser}]{ibrnet}
Wang, Q.; Wang, Z.; Genova, K.; Srinivasan, P.~P.; Zhou, H.; Barron, J.~T.; Martin-Brualla, R.; Snavely, N.; and Funkhouser, T. 2021{\natexlab{b}}.
\newblock Ibrnet: Learning multi-view image-based rendering.
\newblock In \emph{Proceedings of the IEEE/CVF Conference on Computer Vision and Pattern Recognition}, 4690--4699.

\bibitem[{Wang et~al.(2021{\natexlab{c}})Wang, Zhu, Pang, and Lin}]{fcos3d}
Wang, T.; Zhu, X.; Pang, J.; and Lin, D. 2021{\natexlab{c}}.
\newblock Fcos3d: Fully convolutional one-stage monocular 3d object detection.
\newblock In \emph{Proceedings of the IEEE/CVF International Conference on Computer Vision}, 913--922.

\bibitem[{Wang et~al.(2018)Wang, Yu, Huang, and Neumann}]{Sgpn}
Wang, W.; Yu, R.; Huang, Q.; and Neumann, U. 2018.
\newblock Sgpn: Similarity group proposal network for 3d point cloud instance segmentation.
\newblock In \emph{Proceedings of the IEEE conference on computer vision and pattern recognition}, 2569--2578.

\bibitem[{Wang et~al.(2022{\natexlab{b}})Wang, Guizilini, Zhang, Wang, Zhao, and Solomon}]{detr3d}
Wang, Y.; Guizilini, V.~C.; Zhang, T.; Wang, Y.; Zhao, H.; and Solomon, J. 2022{\natexlab{b}}.
\newblock Detr3d: 3d object detection from multi-view images via 3d-to-2d queries.
\newblock In \emph{Conference on Robot Learning}, 180--191. PMLR.

\bibitem[{Xie et~al.(2023)Xie, Jiang, Gkioxari, and Straub}]{PARQ}
Xie, Y.; Jiang, H.; Gkioxari, G.; and Straub, J. 2023.
\newblock Pixel-aligned recurrent queries for multi-view 3d object detection.
\newblock In \emph{Proceedings of the IEEE/CVF International Conference on Computer Vision}, 18370--18380.

\bibitem[{Xu et~al.(2023)Xu, Wu, Hou, Tsai, Li, Wang, Zhan, He, Vajda, Keutzer, and Tomizuka}]{NeRF-Det}
Xu, C.; Wu, B.; Hou, J.; Tsai, S.; Li, R.; Wang, J.; Zhan, W.; He, Z.; Vajda, P.; Keutzer, K.; and Tomizuka, M. 2023.
\newblock NeRF-Det: Learning Geometry-Aware Volumetric Representation for Multi-View 3D Object Detection.
\newblock In \emph{Proceedings of the IEEE/CVF International Conference on Computer Vision (ICCV)}, 23320--23330.

\bibitem[{Yariv et~al.(2021)Yariv, Gu, Kasten, and Lipman}]{volsdf}
Yariv, L.; Gu, J.; Kasten, Y.; and Lipman, Y. 2021.
\newblock Volume rendering of neural implicit surfaces.
\newblock \emph{Advances in Neural Information Processing Systems}, 34: 4805--4815.

\bibitem[{Yu et~al.(2021)Yu, Ye, Tancik, and Kanazawa}]{pixelnerf}
Yu, A.; Ye, V.; Tancik, M.; and Kanazawa, A. 2021.
\newblock pixelnerf: Neural radiance fields from one or few images.
\newblock In \emph{Proceedings of the IEEE/CVF Conference on Computer Vision and Pattern Recognition}, 4578--4587.

\bibitem[{Zhu et~al.(2020)Zhu, Su, Lu, Li, Wang, and Dai}]{deformable}
Zhu, X.; Su, W.; Lu, L.; Li, B.; Wang, X.; and Dai, J. 2020.
\newblock Deformable detr: Deformable transformers for end-to-end object detection.
\newblock \emph{arXiv preprint arXiv:2010.04159}.

\end{thebibliography}
}

\end{document}